\def\year{2018}\relax
\documentclass[letterpaper]{article} 
\usepackage{aaai18}  
\usepackage{times}  
\usepackage{helvet}  
\usepackage{courier}  
\usepackage{url}  
\usepackage{graphicx}  
\usepackage{amsmath}
\usepackage{multirow}
\usepackage{algorithmic}
\usepackage{bm} 

\begin{document}
	%
	\title{Assertion-based QA with Question-Aware Open Information Extraction}
	\author{Zhao Yan$^\S$\thanks{Contribution during internship at \mbox{Microsoft Research Asia}},
		Duyu Tang$^\ddag$, Nan Duan$^\ddag$, Shujie Liu$^\ddag$, Wendi Wang$^\diamond$, \\
		\bf \Large Daxin Jiang$^\diamond$, Ming Zhou$^\ddag$ and Zhoujun Li$^\S$\thanks{Corresponding author}\\
		$^\S$State Key Lab of Software Development Environment, Beihang University, Beijing, China \\
		$^\ddag$Microsoft Research, Beijing, China \ \ \
		$^\diamond$Microsoft, Beijing, China \\		
		\{yanzhao, lizj\}@buaa.edu.cn \ \ \{dutang, nanduan, shujliu, wendw, djiang, mingzhou\}@microsoft.com 
	}

	\maketitle
	\begin{abstract}
		We present assertion based question answering (ABQA), an open domain question answering task that takes a question and a passage as inputs, and outputs a semi-structured assertion consisting of a subject, a predicate and a list of arguments.
		An assertion conveys more evidences than a short answer span in reading comprehension, and it is more concise than a tedious passage in passage-based QA. 
		These advantages make
		ABQA more suitable for human-computer interaction scenarios such as voice-controlled speakers.
		Further progress towards improving ABQA requires richer supervised dataset and powerful models of text understanding.
		To remedy this, we introduce a new dataset called WebAssertions, which includes hand-annotated QA labels for 358,427 assertions in 55,960 web passages. 
		To address ABQA, we develop both generative and extractive approaches.
		The backbone of our generative approach is sequence to sequence learning.
		In order to capture the structure of the output assertion, we introduce a hierarchical decoder that first generates the structure of the assertion and then generates the words of each field.
		The extractive approach is based on learning to rank. Features at different levels of granularity are designed to measure the semantic relevance between a question and an assertion.
		Experimental results show that our approaches have the ability to infer question-aware assertions from a passage.
		We further evaluate our approaches by incorporating the ABQA results as additional features in passage-based QA. 
		Results on two datasets show that ABQA features significantly improve the accuracy on passage-based~QA.
	\end{abstract}

	\section{Introduction}	
	Open-domain question answering (Open-QA) is a long-term goal in natural language processing area, which \mbox{empowers} the computer to answer questions for open domain.
	In this work, we present assertion-based question answering (ABQA), an open QA task that answers a question with semi-structured assertion instead of answer span in machine reading comprehension \cite{squadData2016} or sentence/passage in answer selection \cite{yang2015wikiqa}.
	Here an assertion is a group of words with subject-predicate-object structure which is inferred from the passage guided by the content of the question. We believe that ABQA has many promising advantages.  
	From an industry perspective, ABQA could improve smart speakers such as Amazon Echo, Google Home and Microsoft Invoke, where the scenario is to answer a user's question through reading out a concise and semantically adequate utterance. In this scenario, a short answer span does not convey enough supporting evidences, while a passage is too tedious for a \mbox{speaker}.
	From a research perspective, ABQA is a potential direction to drive explainable question answering. It explicitly reveals the knowledge embodied in the document that answers the question. 
	Moreover, the results from ABQA could be used to improve other QA tasks such as answer sentence selection.
	An assertion graph could also be built on top of these assertions through aggregating the same \mbox{nodes}, which makes explicit reasoning practical \cite{AI2tableILP2016ijcai,AI2OIE4QA2017acl}.
	\begin{table}[t]\small
		\centering
		\begin{tabular}{l|p{0.72\columnwidth}}
			\hline
			\textbf{Question} & who killed jfk \\
			\hline
			\textbf{Method} & \textbf{Answer} \\						
			\hline
			PBQA & A ten-month investigation from November 1963 to September 1964 by the Warren Commission concluded that Kennedy was assassinated by Lee Harvey Oswald, acting alone, and that Jack Ruby also acted alone when he killed Oswald before he could stand trial. \\
			\hline
			MRC & Lee Harvey Oswald \\
			\hline
			ABQA & $<$Kennedy; was assassinated; by Lee \mbox{Harvey Oswald}$>$ \\
			\hline
		\end{tabular}
		\caption{An example to illustrate the difference between three QA tasks, i.e. ABQA, MRC and {PBQA}.}
		\label{Table/DiffMethod}
	\end{table}

	The ABQA task is related to the answer sentence/passage selection (PBQA) task and the machine reading comprehension (MRC) task.
	The difference between ABQA and others is obvious although they all take question-passage pair as the input.
	As Table \ref{Table/DiffMethod} shows, the assertion is organized into a structure with complete and concise information. 
	The ABQA task differs from knowledge based QA (KBQA) in that the knowledge in KBQA is typically curated or extracted from large scale web documents. The goal of ABQA is deep document understanding and to answer question based on that. 
	Various representations of a meaning makes directly linking the knowledge in KB to the document a challenging problem.
	The ABQA task also relates to Open IE (OIE), the goal of which is to extract all the assertions involved in a document. 
	The end goal of ABQA is not only to infer the assertions from both question and document, but also to correctly answer the question.

	To study the ABQA task, we construct a human labeled dataset called WebAssertions.
	The questions and correspond passages are collected from the query log of a commercial search engine in order to reflect the real information need of users.
	For each question-passage pair, we generate assertion candidates by a state-of-the-art OIE algorithm~\cite{clausie}. 
	Human annotators are asked to label whether or not an assertion is correct and concise and meantime can correctly answer the question. 
	The WebAssertions dataset includes hand-annotated QA labels for 358,427 assertions in 55,960 web passages.
	
	We introduce both generative and extractive approaches to address ABQA. 
	Our generative approach which we call Seq2Ast is on the basis of sequence-to-sequence (Seq2Seq) learning. Seq2Ast extends Seq2Seq by incorporating a hierarchical decoder, which first generates the structure of an assertion through a tuple-level decoder, and then generates the words for each slot through another word-level decoder. 
	The extractive method is based on learning to ranking, which ranks candidate assertions with well-designed matching features at different levels.
	
	We conduct experiments on two settings. 
	We first test the performances of our approaches on the ABQA task. 
	\mbox{Results} show that Seq2Ast yields 35.76 in terms of BLEU-4 score, which is better that the Seq2Seq model. 
	We further apply ABQA results as additional features to facilitate the passage-based QA task.
	Results on two datasets \cite{yang2015wikiqa,marcoData2016} show that incorporating ABQA features significantly improve the accuracy on passage-based QA.

	In summary, we make the following contributions:
	\begin{itemize}
		\item We present the ABQA task, which answers a question with an assertion based on the content of a document. We create a manually labeled corpus for ABQA, which will be released to the community.
	
		\item We extend sequence-to-sequence learning approach by introducing a hierarchical decoder to generate the assertion. We also develop an extractive approach for ABQA.

		\item We conduct extensive experiments, and verify the effectiveness of our approach in both ABQA and PBQA tasks.		
	\end{itemize}

	\begin{figure}[t]
		\centering
		\includegraphics[width=0.47\textwidth]
		{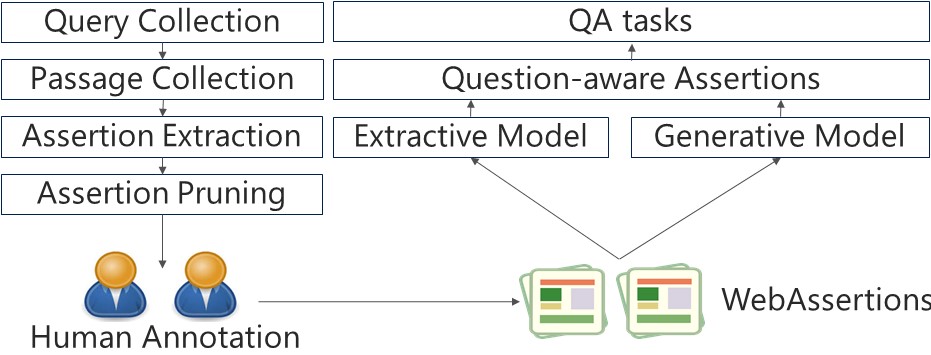}	
		\label{fig/framework}
		\caption{A brief illustration of the dataset construction, ABQA, and the application of ABQA in other QA tasks.}
	\end{figure}

	\section{Task Definition and Dataset Construction} \label{sec/data}
	In this section, we formulate the task of assertion-based question answering (ABQA) and describe the construction of a new dataset tailored for ABQA.
	
	\paragraph{Task Definition} Given a natural language question ($q$) and a passage ($p$), the goal of ABQA is to output a semi-structured assertion ($ast$) that can answer the question $q$ based on the content of the passage $p$.  
	An assertion is represented as an $n$-tuple (n$\geq$3) which consists of a subject ($sub$), a predicate ($pre$), and one or more arguments ($arg_i$). Each field is a natural language sequence that includes one or more words.

	\begin{table}[t]\small
		\centering
		\begin{tabular}{lp{4.9cm}}
			\hline
			
			\textbf{Steps} & \textbf{Details} \\
			\hline
			Query Collection & Collect queries from the search log of a commercial search engine.\\
			Passage Collection & Leverage a search engine, and \mbox{collect} pairs of query-passage if the passage is the direct answer of the query. \\
			Assertion Extraction & Extract candidate assertions from passages based on an open IE algorithm. \\
			Assertion Pruning & Prune assertions based on a combination rule in order to facilitate reasoning. \\
			Human Annotation & Ask labelers to annotate if an assertion can correctly answer the question and meantime has a complete meaning.\\
			\hline		
		\end{tabular}
		\caption{The details of the dataset construction process.}
		\label{table:data-construction}
	\end{table}
	
	\paragraph{Dataset Construction}
	Since there is no publicly available dataset for ABQA, we construct a dataset called WebAssertions through manual annotation. 
	The construction of WebAssertions follows the steps described in Table \ref{table:data-construction}.

	Here we describe some important details during the data construction process.
	There exists several open IE algorithms in literature, including TextRunner \cite{yates2007textrunner}, Reverb \cite{fader2011identifying}, OLLIE \cite{schmitz2012open}. 
	The result of an open IE algorithm has the same format with an assertion.
	We applied these open IE toolkits to a portion of randomly sampled passages from our corpus. 
	We observe that the results extracted via ClausIE answer more questions than other algorithms.
	ClausIE is a rule based open IE algorithm which does not require any training data.
	The backbone of ClausIE is a set of predefined rules, which are based on the structures of sentences obtained via dependency parsing tree. 
	For more details about ClausIE, please refer to \cite{clausie}.

	\begin{table}[h]\small
		\centering
		\begin{tabular}{p{0.07\columnwidth}|c|p{0.66\columnwidth}}
			\hline
			\multicolumn{2}{l|}{\textbf{Question}} & when will shanghai disney open \\
			\hline
			\multicolumn{2}{l|}{\textbf{Passage}} & the Disney empire's latest outpost, Shanghai Disneyland, will open in late 2015, reports the associated press. \\
			\hline
		No. & Label & Assertion \\
			\hline
			1 & 0 & $<$the Disney empire's latest outpost; is; Shanghai Disneyland $>$ \\
			\hline
			2 & 0 & $<$the Disney empire's latest outpost; will open; in late 2015$>$ \\
			\hline
			3 & 0 & $<$the associated press; reports; the Disney empire's latest outpost will open in late 2015$>$ \\
			\hline
			4 & 1 & $<$Shanghai Disneyland; will open; in late 2015 $>$ \\
			\hline
		\end{tabular}
		\caption{A data sample from WebAssertions. The 4$^{th}$ assertion is combined from 1$^{st}$ and 2$^{nd}$ assertion.}
		\label{Table/LabelSample}
	\end{table}
	We use a simple rule to enhance the assertions based on our consideration of facilitating reasoning.
	We believe that ABQA is a great way to drive explainable question answering and reasoning over documents.
	Different from the unexplainable deep neural network approaches in query-passage matching tasks, the structured assertions reveal which portion of knowledge embodied in the document answers the questions.
	Keeping these in mind, in this work we made a preliminary trial to compose new assertions based on the extracted assertions from a document.
	We consider the ``is-a'' relation and use that to do an extension. 
	Supposing two assertions $<$A, $is$, B$>$ and $<$A, $pre$, C$>$ are extracted, we will generate an new assertion $<$B, $pre$, C$>$. 
	An example is given in Table \ref{Table/LabelSample}, in which the 4$^{th}$ assertion is composed based on the 1$^{st}$ and the 2$^{nd}$ assertions.
	Table \ref{Table/LabelSample} gives an example of the human annotation result.
	Data statistics of WebAssertions are given in Table \ref{DataStat}.
	\begin{table}[h]
		\centering
		\begin{tabular}{lc}
			\hline
			\textbf{\# of question-passage} & 55,960 \\
			\textbf{\# of question-assertion} & 358,427 \\
			\textbf{Avg. assertions / question} & 6.41\\
			\textbf{Avg. Words / question} & 6.00\\
			\textbf{Avg. Words / passage} & 39.33\\
			\textbf{Avg. Words / assertion} & 8.62\\
			\hline		
		\end{tabular}
		\caption{Statistics of the WebAssertions.}
		\label{DataStat}
	\end{table}

	\section{Assertion based Question Answering (ABQA)}
	In this section, we describe a generative approach and an extractive approach for ABQA.

	\subsection{Seq2Ast: The Generative Approach for ABQA}
	We develop a sequence-to-assertion (Seq2Ast) approach to generate assertions for ABQA.
	The backbone of Seq2Ast is sequence-to-sequence (Seq2Seq) learning \cite{sutskever2014sequence,cho-EtAl:2014:EMNLP2014}, which has achieved promising performances in a variety of natural language generation tasks. 	
	The Seq2Seq approach includes an encoder and a decoder. The encoder takes a sequence as the input and maps the inputs to a list of hidden vectors. The decoder generates another sequence in a sequential way through outputting a word at one time step.
		
	The main characteristic of ABQA task is that the output in ABQA is an assertion, which is composed of a list of fields and each field consists of a list of words.
	To address this, we present a hierarchical decoder which first generates each field of the assertion through a \textbf{tuple-level} decoder, and then generates the words for each field through a \textbf{word-level} decoder. 
	In Seq2Ast, the tuple-level decoder memorizes the structures of the assertion and the word-level decoder learns the short dependencies in each field.

	Specifically,  we use GRU based RNN \cite{cho-EtAl:2014:EMNLP2014} as the tuple-level decoder to output the representation for each field of the assertion.
	On top of the tuple-based decoder, we use another GRU based RNN as word-level decoder to generate the words of each field.
		\begin{figure}[h]
		\centering
		\includegraphics[width=0.470\textwidth]
		{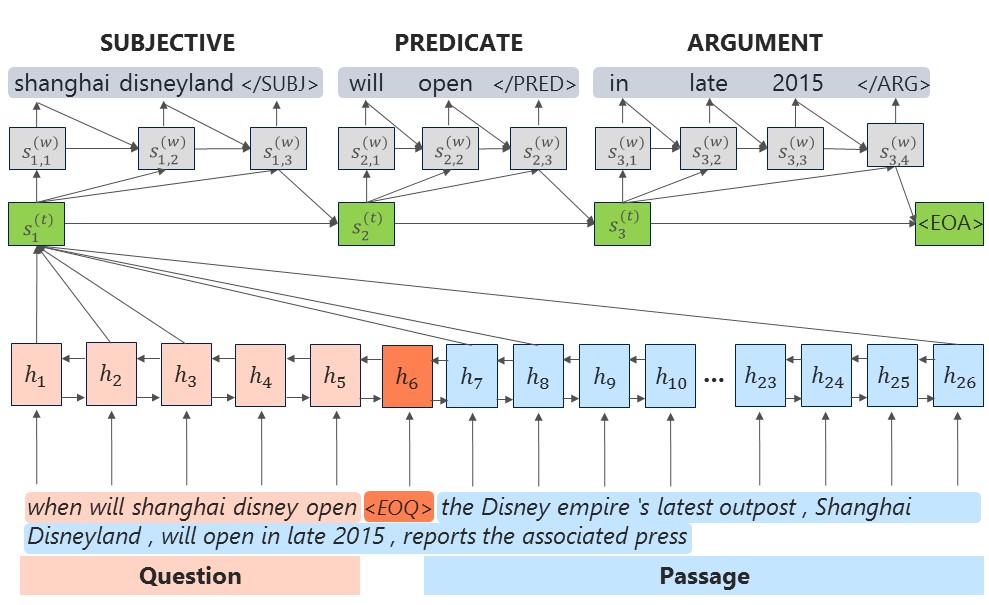}	
		\caption{The architecture of Seq2Ast with a hierarchical decoder.}
		\label{fig/structure}
	\end{figure}
	The architecture of Seq2Ast is given in Figure \ref{fig/structure}, which is inspired by the chunk-based NMT \cite{ishiwatari-EtAl:2017:Long}.
		To generate the representation of the $k$-th field ${{\bm s}}^{(t)}_k$, the tuple-level decoder takes the last state of the word-level decoder ${\bm s}^{(w)}_{k-1, J_{k-1}}$ and updates its hidden state ${\bm s}^{(t)}_{k}$ as follows:
	\begin{align}
	{\bm s}^{(t)}_{k} &= \mbox{\textsc{gru}}({\bm s}^{(t)}_{k-1}, {\bm s}^{(w)}_{k-1, J_{k-1}})  
	\end{align}

	We consider the field representation as a global information to guide the prediction of each word, therefore the field representation ${{\bm s}}^{(t)}_{k}$ is also fed to the word-level decoder as additional input until it outputs all the words in the current field.	
	Attention \cite{bahdanau2014neural} is used in the word level decoder in order to selectively retrieve important content from the encoder part. 
	To deal with the rare word problem, we use a simple yet effective copying mechanism which replaces the generated out-of-vocabulary word with a word of largest attention score from the source.
	\begin{align}
	&{\bm s}^{(w)}_{k,j} = \mbox{\textsc{gru}}({\bm s}^{(w)}_{k,j-1}, [{{\bm s}}^{(t)}_{k}; {\bm y}_{k,j-1}])  
	\end{align}

	We use bidirectional GRU based RNN as the encoder.
	In this work, we concatenate the passage and the question, separated by a special tag.\footnote{
		An alternative way is to regard the document as the memory \cite{Sukhbaatar2015e2emn} and use the question to iteratively retrieve from and update the memory. 
		In this paper, we favor to the simple concatenation strategy.} 
	The model is learned in an end-to-end way with back-propagation, the objective of which is to maximize the probability of the correct assertion given a question-passage pair.
	In the experiment, the parameters in Seq2Ast are randomly initialized, and updated with AdeDelta \cite{zeiler2012adadelta}.

	\subsection{ExtAst: The Extractive method for ABQA}
	The extractive method is a learning to rank based method which selects top-ranked assertion from a candidate list based on features designed in different granularities.
	Our extractive method includes three steps: 
	i) assertion candidate generation which has been described in the dataset construction process; 
	ii) question-aware matching features; 
	iii) assertion candidate ranking.

	\subsubsection{Question-aware Matching Features}
	We design features at three different levels of granularities to measure the semantic relevance between a question ($q$) and an assertion ($ast$).
	
	In \textbf{Word-Level}, we use a word matching feature $f_{WM}$ and a word level translation feature $f_{W2W}$.	
	The intuition of $f_{WM}$ is that an assertion is relevant to a question if they have a large amount of word overlap.
	The $f_{WM}$ are calculated based on number of words shared by question and assertion.
	The $f_{W2W}$ denotes a word-to-word translation-based feature that calculates the relevance for a question and an assertion based on IBM model 1~\cite{brown1993IBM1}.
	The probabilities of word alignments are trained on 11.6M ``sentence-similar sentence'' pairs by GIZA++~\cite{och2003GIZApp}.

	In \textbf{Phrase-Level},
	we design a paraphrase-based feature $f_{PP}$ and a phrase-to-phrase translation feature $f_{P2P}$ to deal with the case that a question and an assertion use different expressions to describe the same meaning.	
	Both $f_{PP}$ and $f_{P2P}$ are based on extracted phrase tables (PT) by existing statistical machine translation method~\cite{koehn2003PP}.
	The difference between $f_{PP}$ and $f_{P2P}$ is that the PT of 
	$f_{PP}$ is extracted from 0.5M ``English-Chinese'' bilingual pairs while the PT of $f_{P2P}$ is extracted from 4M ``question-answer'' pairs.

	In \textbf{Sentence-Level},	
	we use a CNN-based feature $f_{CNN}$ and a RNN-based feature $f_{RNN}$ to match a question to an assertion.
	The $f_{CNN}$ features is based on the CDSSM model~\cite{shen2014CDSSM}, a convolutional neural network approach which has been successfully applied in sentence matching tasks.
	The model composes question vector and assertion vector via two convolutional neural networks separately and calculate their relevance with cosine function.
	\begin{eqnarray}
		f_{CNN}(que, ast)=cosine(cdssm_{1}(q), cdssm_{2}(ast)) 
	\end{eqnarray}

	We also use a recurrent neural network based model to calculate $f_{RNN}$.
	We first use two RNNs to map a question and an assertion to fixed-length vectors separately.
	The same bi-directional GRU is used to get the question representation and assertion representation from both directions.
	Taking question representation as an example, 
	it recursively transforms current word vector $e_{q, t}$ with the output vector of the previous step $h_{q, t-1}$.	
	In the representation layer, we concatenate the four last hidden states and the element-wise multiplication of the vectors from both directions as the final representation.
	Afterwards, we feed the representation of question-assertion pair to a multilayer perceptron (MLP).
		
	We train model parameters of $f_{CNN}$ and $f_{RNN}$ on 4M ``question-answer'' pairs with stochastic gradient descent. 
	The pair-wise margin ranking loss for each training instance is calculated as: 
	\begin{eqnarray}
	&\mathcal{L} &=\max(0, m-f^{+}(q, ast)+f^{-}(q, ast)), 
	\end{eqnarray} 
	where $f^{+}(q, ast)$ and $f^{-}(q, ast)$ are the model scores for a relevance and irrelevance pair and $m$ is the margin.
	
	\subsubsection{Assertion Candidate Ranking}	
	We use LambdaMART \cite{mart2010l2r}, an algorithm for solving real world ranking problem, to learn the final ranking score of each question-assertion pair.\footnote{We also implemented a ranker with logistic regression, however, its performance was obviously worse than LambdaMART in our experiment.}
	The basic idea of LambdaMART is that it constructs a forest of decision trees, and its output is a linear combination of the results of decision trees.
	Each branch in a decision tree specifies a threshold to apply to a single feature, and each leaf node is a real value.
	Specifically, for a forest of $N$ trees, the relevance score of a question-assertion pair is calculated as 
	\begin{equation}
	s(q, ast) = \sum_{i=1}^{N} w_i tr_i(q, ast),   
	\end{equation}
	where $w_i$ is the weight associated with the $i$-th regression tree, and $tr_i( \cdot )$ is the value of a leaf node obtained by evaluating $i$-th tree with features $\left[ f_1(q, ast), ... ,f_K(q, ast) \right]$.
	The values of $w_i$ and the parameters in $tr_i(\cdot)$ are learned with gradient descent during training.

	\section{Experiment}
	\begin{table*}[t]
		\centering
		\begin{tabular}{p{0.1\columnwidth}|c|p{1.65\columnwidth}}	
			\hline
			\multicolumn{2}{l|}{\textbf{Question}} & how much can your bladder hold
			\\
			\hline
			\multicolumn{2}{l|}{\textbf{Passage}} & A healthy adult bladder can hold up to 16 ounces (2 cups) of urine comfortably, according to the national institutes of health. How frequently it fills depends on how much excess water your body is trying to get rid of.
			\\
			\hline
			\multicolumn{2}{l|}{\textbf{Generative Result}} & $<$a healthy adult bladder; can hold; up to 16 ounces$>$ \\
			\hline
			\multicolumn{3}{l}{\textbf{Extractive Result}} \\
			\hline
			\textbf{Rank} & \textbf{Label} & \textbf{Assertion} \\
			\hline
			\multicolumn{1}{c|}{1} & 1 & $<$a healthy adult bladder; can hold; up to 16 ounces; 2 cups of urine$>$ \\
			\hline
			\multicolumn{1}{c|}{2} & 0 & $<$a healthy adult bladder; can hold; up to 16 ounces; according to the national institutes of health$>$ \\
			\hline
			\multicolumn{1}{c|}{3} & 0 & $<$a healthy adult bladder; can hold; up to 16 ounces; comfortably$>$ \\
			\hline
			\multicolumn{1}{c|}{4} & 0 & $<$it; fills; how frequently$>$ \\
			\hline
			\multicolumn{1}{c|}{5} & 0 & $<$your body; is trying; to rid of; how much excess water$>$ \\
			\hline
		\end{tabular}
		\caption{An example illustrating the results of the generative and the extractive approaches.}
		\label{Table/CaseStudy}
	\end{table*}
	In this section, we describe experimental settings and report empirical results on ABQA and the application of ABQA in the answer sentence selection task.
	\subsection{Results on ABQA}
	We first test the generative and extractive approaches on the assertion-based question answering (ABQA) task.
	In this experiment, we randomly split the WebAssertions dataset into training, development, and test sets with a 80:10:10 split. 
	Parameters are tuned on the development set and results are reported on the test set. The test set contains 36,165 question-passage-assertion triples from 5,575 question-passage pairs.

	We first conduct evaluation from a text generation perspective. We use BLEU-4 score \cite{papineni2002bleu} as the automatic evaluation metric, the goal of which is to measure the ngram match between the generated assertion and the referenced assertion. We compare to the standard Seq2Seq model w/ and w/o attention mechanism. Results are given in Table \ref{bleu}. We can see that Seq2Ast performs better than the standard Seq2Seq method, which verifies the effectiveness of the hierarchical decoder.
	As a reference, we can also report the BLEU-4 score of the extractive approach despite this is not a perfect to compare between the generative and extractive approaches. The BLUE-4 score of our extractive approach is 72.27, which is extremely high for a text generation task. But this is also reasonable because the extractive approach aims to select a most possible assertion from a candidate list which includes the referenced result. Therefore, the BLEU-4 score for a correct top-ranked result is 100. Further experiments that applying the results of both generative and extractive approaches in passage-level question answering task will be given in the following subsection.
	
	\begin{table}[h]
		\centering
		\begin{tabular}{cc}
			\hline
			\textbf{Method} &  \textbf{BLEU-4} \\	
			\hline
			Seq2Seq	& 22.01 \\	
			Seq2Seq	+ attention & 31.85 \\		
			Seq2Ast  & 35.76 \\
			\hline
		\end{tabular}
		\caption{Performance on generative based ABQA.}
		\label{bleu}
	\end{table}
	
	We evaluate our extractive method as a ranking problem, the goal of which is to rank the assertion candidates for a given question-passage pair and select the assertion that has the largest probability to correctly  answer the question. 
	Hence, we choose Precision@1 (P@1), Mean Average Precision (MAP) and Mean Reciprocal Rank (MRR) to evaluate the performance of our model \cite{manning2008ir}.

	We conduct an ablation test to study the effects of different features in the extractive approach.  
	Results are given in Table \ref{EvalAssRank}.
	It is not surprising that sentence-level feature performs better than word-level and phrase-level features because of better modeling the global semantic relevance between a question and an assertion.
	Our system ExtAst that combines all the features obtains the best performance. 
	
	\begin{table}[h]
		\centering
		\begin{tabular}{cccc}
			\hline
			\textbf{Methods} & \textbf{MAP} & \textbf{MRR} & \textbf{P@1}\\                    
			\hline
			WordMatch & 65.85\% & 66.67\% & 47.62\% \\
			Word-Level & 71.13\% & 72.08\% & 55.47\%\\
			Phrase-Level & 72.18\% & 72.86\% & 56.74\%\\
			Sentence-Level & 76.49\% & 77.45\% & 63.34\%\\
			ExtAst & 77.99\% & 78.90\% & 65.56\% \\
			\hline
		\end{tabular}
		\caption{Performances on extractive based ABQA.}
		\label{EvalAssRank}
	\end{table}
	
	\begin{table*}[t]
		\centering
		\begin{tabular}{cc|cc|cc}
			\hline
			& \multirow{2}{*}{\textbf{Methods}} & \multicolumn{2}{c|}{WikiQA} & \multicolumn{2}{c}{MARCO} \\
			\cline{3-6}
			 &  & \textbf{MAP} & \textbf{MRR} & \textbf{MAP} & \textbf{MRR}\\
			\hline
			\multicolumn{2}{l|}{\textbf{Published Models}} &&&\\
			(1) & CNN+Cnt~\cite{yang2015wikiqa} & 65.20\% & 66.52\% & - & - \\
			(2) & LSTM+Att+Cnt~\cite{miao2015NASM} & 68.55\% & 70.41\% & - & - \\ 
			(3) & ABCNN~\cite{yin2015abcnn} & 69.21\% & 71.08\% & 46.91\% & 47.67\% \\
			(4) & Dual-QA~\cite{tang2017question} & 68.44\% & 70.02\% & 48.36\% & 49.11\%\\
			(5) & IARNN-Occam~\cite{wang2016inner} & 73.41\% & 74.18\% & - & -\\
			(6) & conv-RNN~\cite{wang2017hybrid} & \textbf{74.27}\% & 75.04\% & - & -\\
			(7) & CNN+CH~\cite{tymoshenko2016convolutional} & 73.69\% & \textbf{75.88}\% & - & - \\ 
			\hline
			\multicolumn{2}{l|}{\textbf{Our Models}} &&&\\
			(8) & Baseline & 69.89\% & 71.33\% & 45.97\% & 46.62\%\\
			(9) & Baseline+RndAst & 69.17\% & 70.12\% & 46.62\% & 47.27\% \\ 
			(10) & Baseline+MaxAst & 71.82\% & 72.81\% & 49.37\% & 50.05\% \\ 
			(11) & Baseline+ExtAst & 72.33\% & 73.52\% &\textbf{50.07\%} & \textbf{50.76\%}\\
			(12) & Baseline+Seq2Ast & 72.26\% & 73.35\% & 47.44\% & 48.10\% \\ 
			\hline
		\end{tabular}
		\caption{Evaluation of answer selection task on WikiQA and MARCO datasets.}
		\label{SenSel}
	\end{table*}
	
	A sampled instance together with the results of our generative and extractive approaches are illustrated in Table \ref{Table/CaseStudy}.
	We can see that the generative model has the ability to produce the structure of an assertion, fluent expressions for each field of the assertion and a complete meaning to some extent. In this example, the generative result is even better than the extractive model in terms of concise.
	However, the generative model is far more perfect. After doing case studies, we find that fluency is not a big issue. 
	The main issue of current approach includes generating duplicate content and generating an assertion that is irrelevant to the question.
	The first issue would be mitigated with coverage mechanism \cite{tu-EtAl:2016:P16-1}, which explicitly memorizes whether a source word has been replicated or not. 
	Addressing the second issue might require exploring deep question understanding, and a decoder that is deeply driven by the question.
	
	We also conduct error analysis on the extractive approach. 
	We summarize the main errors into three categories. The first category is question type mismatch. 
	For instance, the answer for ``\textit{When were the Mongols defeated by the Tran?}'' is a reasonable assertion yet does not contain any time information.
	The second category is the mismatch between the entity in query and the different expressions in the passage. Co-reference resolution could be also grouped into this category.
	The third category is the require of reasoning. An exampled question is ``\textit{Which is the largest city not connected to an interstate highway?}''. Our current model does not have the ability to handle the  ``not'' type question.

	\subsection{Improve PBQA with ABQA results}	
	We further evaluate the performance of our ABQA algorithms by applying the results into the passage-based question answering task (PBQA), and use the end-to-end performance in PBQA to reflect the effects of our approaches.
	In this work, we use answer selection as the PBQA tasks, which takes a question and a passage as the input, and outputs a sentence which comes from the passage as the answer. 
	
	Given a question and a document, we first use our ABQA algorithms to output the top-ranked assertion through generative or extractive approaches. Afterwards, additional features for a question-assertion pair is appended to the original feature vector which is used for answer sentence selection. We use exactly the same feature set which we have used in the extractive ABQA approach.
	The basis features for answer sentence selection include a word-level feature based on the number of occurred words in both question and passage,  and a sentence-level feature that encodes both question and passage as continuous vector with convolutional neural network. 
	We also employ LambdaMART to train the ranking model for answer sentence selection.
	Feature weights in the ranking model are trained by SGD based on the training data that consists of a set of labeled $<$ question, sentence, label$>$ triples,
	where $label$ indicates whether the sentence is the correct answer for the question or not.

	Results are reported on WikiQA and MARCO datasets, both of which are suitable to test our ABQA approach as the questions from these dataset are also real user queries from the search engine, which is consistent with the WebAssertions dataset. 
	WikiQA is a benchmark dataset for answer sentence selection and precisely constructed based on natural language questions and Wikipedia documents.
	WikiQA dataset contains 20,360 instances in the training set,  2,733 instances in development set, and 6,165 instances in the test set.
	The MARCO dataset is originally constructed for the reading comprehension task, yet also includes manual annotation for passage ranking.
	In MARCO dataset,  \mbox{questions} come from Bing search log and passage candidates come from search engine's results. 
	Annotators will label a passage as 1 if the passage contains evidences to answer the given question. 
	Since the ground truth of the MARCO's test set is invisible to public, we randomly split the original validation set into dev set and test set.
	In this paper we only use the information about passage selection to test our model.
	The MARCO dataset contains 676,193 instances in the training set, 39,510 instances in the development set, and 42,850 instances in the test set. 
	In this experiment, we also use MAP and MRR as the evaluation metrics.
	Similar to other published works, the  calculation of the evaluation metric does not include the instances whose candidate answers are all incorrect or all correct .

	We compare to different algorithms for PBQA.
	\mbox{Results} are given in Table \ref{SenSel}.
	The results of baseline approaches on these two datasets are reported in previous \mbox{publications.} 
	\textbf{\mbox{CNN}+Cnt}~\cite{yang2015wikiqa} combines a bi-gram CNN model with word count by logistic regression. 
	\textbf{\mbox{LSTM}+Att+Cnt}~\cite{miao2015NASM}  combines an attention-based LSTM model with word count by logistic regression.
	\textbf{ABCNN}~\cite{yin2015abcnn} uses an attention-based CNN model which has been proven very powerful in various sentence matching tasks. 
	\textbf{Dual-QA}~\cite{tang2017question} take QA and question generation (QG) as dual task.
	The result of ABCNN model on MARCO dataset is reported in~\cite{tang2017question}.
	\textbf{IARNN-Occam}~\cite{wang2016inner} is a RNN model with inner attention mechanism.
	\textbf{conv-RNN}~\cite{wang2017hybrid} is an hybrid model that combines both CNN and RNN.
	\textbf{CNN+CH}~\cite{tymoshenko2016convolutional} is an hybrid model combined convolution tree kernel features with \mbox{CNN}.
	As described before, our baseline system contains a word-level feature based on word overlap and a sentence-level feature based on CDSSM~\cite{shen2014CDSSM}.

	We further compare to different usages of assertions for PBQA. 
	Without using our question-aware assertion generation/extraction approach, we could also use open IE approaches to extract all the assertions from the passage, and then aggregate these assertions as additional features for \mbox{PBQA}.
	We implement two strategies towards this goal. The \textbf{RndAst} means that we randomly select an assertion and use it to calculate the additional assertion-level feature vector.
	The \textbf{MaxAst} is similar to the max-pooling operation in convolutional neural network. We first get the feature vectors for all the extracted assertions from a passage, and then select the max value in each dimension from a list of feature vectors.
	From the results, we can see that our approaches (especially ExtAst) significantly improves our baseline system.

	\section{Related Work}
	Our work relates to the fields of open information extraction, open knowledge-based QA, passage-based QA and machine reading comprehension.

	The ABQA task is related to the Machine Reading Comprehension (MRC) \cite{squadData2016} task in that both take question-passage pair as the input.	
	The difference between ABQA and MRC is that the output of ABQA is an assertion which organized as a semi-structure with complete and concise information, while the output of MRC is a short answer span.
	The ABQA task also differs from passage based QA (PBQA) where the answer is a long passage. 
	Our extractive method is related to existing works for \mbox{PBQA}.			 
	\mbox{LCLR}~\cite{Yih2013LCLR} applied rich lexical semantic features obtained from a wide range of linguistic resources including WordNet, polarity-inducing \mbox{latent} semantic analysis (PILSA) model and different vector \mbox{space} models.
	Convolutional neural \mbox{networks}~\cite{yu2014bicnn,severyn2015cnnRank} and recurrent neural \mbox{networks}~\cite{wang2015lstm} are used to encode questions and answer passages into a semantic vector space.
	ABCNN~\cite{yin2015abcnn} is an attention based CNN which first calculates an similarity matrix and takes it as a new channel of the \mbox{CNN} model.
	Recent studies~\cite{duan-EtAl:2017:EMNLP2017,tang2017question} also explore question generation to improve question answering system. 
	
	Open IE works extract triples of format $<$subject, predicate, arguments$>$ from text in natural language, and does not presuppose a predefined set of predicates.
    TextRunner~\cite{yates2007textrunner} is a pioneering Open IE work which aims at constructing a general model that expresses a relation
	based on Part-of-Speech and Chunking features. 		
	ReVerb~\cite{fader2011identifying} restricts the predicates to verbal phrases and extracts them based on grammatical structures.  
	ClausIE~\cite{clausie} employs hand-crafted grammatical patterns based on the result of dependency parsing trees to detect and extract clause-based assertions. 
	This work differs from Open IE in that the end goal of our work is not only to infer the assertions from both question and document, but also to correctly answer the question.
	In addition, our generative method has the ability to generate words that do not occur in the source text.
	
	There are two lines of studies in knowledge-based question answering (KBQA). 
	One focuses on answering natural language question with curated KB \cite{berant2013semantic,bao2014kbqa,Yih2015acl}, where the key problem is how to link questions in natural language to the structured knowledge in KB. 
	Another line of research in KBQA focuses on large-scale open KB which is automatically extracted from web corpora by means of open IE techniques. 
	To address KBQA, Fader \textit{et al.}~\cite{fader2013paraphrase} present the first open KBQA system which learns question paraphrases over a large corpus.
	OQA \cite{fader2014OQA} is a system that processes questions using a cascaded pipeline on both curated and open KBs.
	TAQA \cite{yin2015answering} is an open KBQA system, which operates on $n$-tuple assertions in order to answer questions with complex semantic constraints.
	TUPLEINF~\cite{AI2OIE4QA2017acl} answers complex questions by reasoning over Open IE knowledge with an integer linear programming (ILP) optimization model, and searches for the best subset of assertions.
	ABQA differs from KBQA in that the assertions/knowledge are extracted from the document, and the focus of ABQA is \mbox{document} understanding and answering question based on that. 
	Our method also differs from KBQA works in that the knowledge in KBQA is typically curated or extracted from large scale web documents beforehand, while our goal is to infer knowledge based on the question and the document.

	\section{Conclusion}
	
	In this paper, we introduce assertion-based question answering (ABQA), an open-domain QA task that answers a question with a semi-structured assertion which is inferred (generated or extracted) from the content of a document.
	We construct a dataset called WebAssertions tailored for ABQA and develop both generative and extractive approaches.
	We conduct extensive experiments in various settings. 
	\mbox{Results} show that our ABQA approaches have the ability to infer question-aware assertions from the document.
	We also demonstrate that incorporating ABQA results as additional features significantly improves the accuracy of a baseline system on passage-based QA. We plan to improve the question understanding component and the reasoning ability of the approach so that  assertions across different sentences could be used to infer the final answer.
	
	\section{ Acknowledgments}
	We greatly thank the anonymous reviews for their valuable comments.
	This work is supported by the National Natural Science Foundation of China (Grand Nos. 61672081, U1636211, 61370126),
	Beijing Advanced Innovation Center for Imaging Technology (No.BAICIT-2016001),
	National High Technology Research and Development Program of China under grant (No.2015AA016004).

	\bibliographystyle{aaai}
	\bibliography{aaai18}

\begin{thebibliography}{}

\bibitem[\protect\citeauthoryear{Bahdanau, Cho, and
  Bengio}{2014}]{bahdanau2014neural}
Bahdanau, D.; Cho, K.; and Bengio, Y.
\newblock 2014.
\newblock Neural machine translation by jointly learning to align and
  translate.
\newblock {\em arXiv preprint arXiv:1409.0473}.

\bibitem[\protect\citeauthoryear{Bao \bgroup et al\mbox.\egroup
  }{2014}]{bao2014kbqa}
Bao, J.; Duan, N.; Zhou, M.; and Zhao, T.
\newblock 2014.
\newblock Knowledge-based question answering as machine translation.
\newblock {\em Proceedings of ACL} 2:6.

\bibitem[\protect\citeauthoryear{Berant \bgroup et al\mbox.\egroup
  }{2013}]{berant2013semantic}
Berant, J.; Chou, A.; Frostig, R.; and Liang, P.
\newblock 2013.
\newblock Semantic parsing on freebase from question-answer pairs.
\newblock In {\em Proceedings of 2013 EMNLP}, volume~2, ~6.

\bibitem[\protect\citeauthoryear{Brown \bgroup et al\mbox.\egroup
  }{1993}]{brown1993IBM1}
Brown, P.~F.; Pietra, V. J.~D.; Pietra, S. A.~D.; and Mercer, R.~L.
\newblock 1993.
\newblock The mathematics of statistical machine translation: Parameter
  estimation.
\newblock {\em Computational linguistics} 19(2):263--311.

\bibitem[\protect\citeauthoryear{Burges}{2010}]{mart2010l2r}
Burges, C.~J.
\newblock 2010.
\newblock From ranknet to lambdarank to lambdamart: An overview.
\newblock {\em Microsoft Research Technical Report MSR-TR-2010-82}
  11(23-581):81.

\bibitem[\protect\citeauthoryear{Cho \bgroup et al\mbox.\egroup
  }{2014}]{cho-EtAl:2014:EMNLP2014}
Cho, K.; van Merrienboer, B.; Gulcehre, C.; Bahdanau, D.; Bougares, F.;
  Schwenk, H.; and Bengio, Y.
\newblock 2014.
\newblock Learning phrase representations using rnn encoder--decoder for
  statistical machine translation.
\newblock In {\em Proceedings of the EMNLP},  1724--1734.

\bibitem[\protect\citeauthoryear{Del~Corro and Gemulla}{2013}]{clausie}
Del~Corro, L., and Gemulla, R.
\newblock 2013.
\newblock Clausie: clause-based open information extraction.
\newblock In {\em Proceedings of the 22nd international conference on WWW},
  355--366.

\bibitem[\protect\citeauthoryear{Duan \bgroup et al\mbox.\egroup
  }{2017}]{duan-EtAl:2017:EMNLP2017}
Duan, N.; Tang, D.; Chen, P.; and Zhou, M.
\newblock 2017.
\newblock Question generation for question answering.
\newblock In {\em Proceedings of the 2017 Conference on Empirical Methods in
  Natural Language Processing},  877--885.
\newblock Association for Computational Linguistics.

\bibitem[\protect\citeauthoryear{Fader, Soderland, and
  Etzioni}{2011}]{fader2011identifying}
Fader, A.; Soderland, S.; and Etzioni, O.
\newblock 2011.
\newblock Identifying relations for open information extraction.
\newblock In {\em Proceedings of the Conference on EMNLP},  1535--1545.

\bibitem[\protect\citeauthoryear{Fader, Zettlemoyer, and
  Etzioni}{2013}]{fader2013paraphrase}
Fader, A.; Zettlemoyer, L.~S.; and Etzioni, O.
\newblock 2013.
\newblock Paraphrase-driven learning for open question answering.
\newblock In {\em ACL (1)},  1608--1618.

\bibitem[\protect\citeauthoryear{Fader, Zettlemoyer, and
  Etzioni}{2014}]{fader2014OQA}
Fader, A.; Zettlemoyer, L.; and Etzioni, O.
\newblock 2014.
\newblock Open question answering over curated and extracted knowledge bases.
\newblock In {\em Proceedings of the 20th ACM SIGKDD},  1156--1165.
\newblock ACM.

\bibitem[\protect\citeauthoryear{Ishiwatari \bgroup et al\mbox.\egroup
  }{2017}]{ishiwatari-EtAl:2017:Long}
Ishiwatari, S.; Yao, J.; Liu, S.; Li, M.; Zhou, M.; Yoshinaga, N.; Kitsuregawa,
  M.; and Jia, W.
\newblock 2017.
\newblock Chunk-based decoder for neural machine translation.
\newblock In {\em Proceedings of the 55th ACL},  1901--1912.

\bibitem[\protect\citeauthoryear{Khashabi \bgroup et al\mbox.\egroup
  }{2016}]{AI2tableILP2016ijcai}
Khashabi, D.; Khot, T.; Sabharwal, A.; Clark, P.; Etzioni, O.; and Roth, D.
\newblock 2016.
\newblock Question answering via integer programming over semi-structured
  knowledge.
\newblock {\em Proceedings of the IJCAI-16}  1145--1152.

\bibitem[\protect\citeauthoryear{Khot, Sabharwal, and
  Clark}{2017}]{AI2OIE4QA2017acl}
Khot, T.; Sabharwal, A.; and Clark, P.
\newblock 2017.
\newblock Answering complex questions using open information extraction.
\newblock In {\em Proceedings of the 55th ACL},  311--316.

\bibitem[\protect\citeauthoryear{Koehn, Och, and Marcu}{2003}]{koehn2003PP}
Koehn, P.; Och, F.~J.; and Marcu, D.
\newblock 2003.
\newblock Statistical phrase-based translation.
\newblock {\em Proceedings of Annual Conference of the (NAACL-HLT)} 1:48--54.

\bibitem[\protect\citeauthoryear{Manning \bgroup et al\mbox.\egroup
  }{2008}]{manning2008ir}
Manning, C.~D.; Raghavan, P.; Sch{\"u}tze, H.; et~al.
\newblock 2008.
\newblock {\em Introduction to information retrieval}, volume~1.
\newblock Cambridge university press Cambridge.

\bibitem[\protect\citeauthoryear{Miao, Yu, and Blunsom}{2015}]{miao2015NASM}
Miao, Y.; Yu, L.; and Blunsom, P.
\newblock 2015.
\newblock Neural variational inference for text processing.
\newblock {\em arXiv preprint arXiv:1511.06038}.

\bibitem[\protect\citeauthoryear{Nguyen \bgroup et al\mbox.\egroup
  }{2016}]{marcoData2016}
Nguyen, T.; Rosenberg, M.; Song, X.; Gao, J.; Tiwary, S.; Majumder, R.; and
  Deng, L.
\newblock 2016.
\newblock Ms marco: A human generated machine reading comprehension dataset.
\newblock {\em arXiv preprint arXiv:1611.09268}.

\bibitem[\protect\citeauthoryear{Och and Ney}{2003}]{och2003GIZApp}
Och, F.~J., and Ney, H.
\newblock 2003.
\newblock A systematic comparison of various statistical alignment models.
\newblock {\em Computational Linguistics} 29(1):19--51.

\bibitem[\protect\citeauthoryear{Papineni \bgroup et al\mbox.\egroup
  }{2002}]{papineni2002bleu}
Papineni, K.; Roukos, S.; Ward, T.; and Zhu, W.-J.
\newblock 2002.
\newblock Bleu: a method for automatic evaluation of machine translation.
\newblock In {\em Proceedings of Annual Meeting of the Association for
  Computational Linguistics (ACL)},  311--318.

\bibitem[\protect\citeauthoryear{Rajpurkar \bgroup et al\mbox.\egroup
  }{2016}]{squadData2016}
Rajpurkar, P.; Zhang, J.; Lopyrev, K.; and Liang, P.
\newblock 2016.
\newblock Squad: 100,000+ questions for machine comprehension of text.
\newblock In {\em Proceedings of the 2016 Conference on Empirical Methods in
  Natural Language Processing},  2383--2392.

\bibitem[\protect\citeauthoryear{Schmitz \bgroup et al\mbox.\egroup
  }{2012}]{schmitz2012open}
Schmitz, M.; Bart, R.; Soderland, S.; Etzioni, O.; et~al.
\newblock 2012.
\newblock Open language learning for information extraction.
\newblock In {\em Proceedings of the EMNLP},  523--534.

\bibitem[\protect\citeauthoryear{Severyn and
  Moschitti}{2015}]{severyn2015cnnRank}
Severyn, A., and Moschitti, A.
\newblock 2015.
\newblock Learning to rank short text pairs with convolutional deep neural
  networks.
\newblock In {\em Proceedings of ACM SIGIR},  373--382.

\bibitem[\protect\citeauthoryear{Shen \bgroup et al\mbox.\egroup
  }{2014}]{shen2014CDSSM}
Shen, Y.; He, X.; Gao, J.; Deng, L.; and Mesnil, G.
\newblock 2014.
\newblock A latent semantic model with convolutional-pooling structure for
  information retrieval.
\newblock In {\em Proceedings of the Conference on Information and Knowledge
  Management},  101--110.

\bibitem[\protect\citeauthoryear{Sukhbaatar \bgroup et al\mbox.\egroup
  }{2015}]{Sukhbaatar2015e2emn}
Sukhbaatar, S.; Szlam, A.; Weston, J.; and Fergus, R.
\newblock 2015.
\newblock End-to-end memory networks.
\newblock In {\em Advances in Neural Information Processing Systems (NIPS)},
  2431--2439.

\bibitem[\protect\citeauthoryear{Sutskever, Vinyals, and
  Le}{2014}]{sutskever2014sequence}
Sutskever, I.; Vinyals, O.; and Le, Q.~V.
\newblock 2014.
\newblock Sequence to sequence learning with neural networks.
\newblock In {\em Advances in neural information processing systems},
  3104--3112.

\bibitem[\protect\citeauthoryear{Tang \bgroup et al\mbox.\egroup
  }{2017}]{tang2017question}
Tang, D.; Duan, N.; Qin, T.; and Zhou, M.
\newblock 2017.
\newblock Question answering and question generation as dual tasks.
\newblock {\em arXiv preprint arXiv:1706.02027}.

\bibitem[\protect\citeauthoryear{Tu \bgroup et al\mbox.\egroup
  }{2016}]{tu-EtAl:2016:P16-1}
Tu, Z.; Lu, Z.; Liu, Y.; Liu, X.; and Li, H.
\newblock 2016.
\newblock Modeling coverage for neural machine translation.
\newblock In {\em Proceedings of the 54th ACL},  76--85.

\bibitem[\protect\citeauthoryear{Tymoshenko, Bonadiman, and
  Moschitti}{2016}]{tymoshenko2016convolutional}
Tymoshenko, K.; Bonadiman, D.; and Moschitti, A.
\newblock 2016.
\newblock Convolutional neural networks vs. convolution kernels: Feature
  engineering for answer sentence reranking.
\newblock In {\em HLT-NAACL},  1268--1278.

\bibitem[\protect\citeauthoryear{Wang and Nyberg}{2015}]{wang2015lstm}
Wang, D., and Nyberg, E.
\newblock 2015.
\newblock A long short-term memory model for answer sentence selection in
  question answering.
\newblock In {\em Proceedings of the 53nd ACL},  707--712.

\bibitem[\protect\citeauthoryear{Wang, Jiang, and Yang}{2017}]{wang2017hybrid}
Wang, C.; Jiang, F.; and Yang, H.
\newblock 2017.
\newblock A hybrid framework for text modeling with convolutional rnn.
\newblock In {\em Proceedings of the 23rd ACM SIGKDD},  2061--2069.
\newblock ACM.

\bibitem[\protect\citeauthoryear{Wang, Liu, and Zhao}{2016}]{wang2016inner}
Wang, B.; Liu, K.; and Zhao, J.
\newblock 2016.
\newblock Inner attention based recurrent neural networks for answer selection.
\newblock In {\em Proceedings of the 54th ACL}.

\bibitem[\protect\citeauthoryear{Yang, Yih, and Meek}{2015}]{yang2015wikiqa}
Yang, Y.; Yih, W.-t.; and Meek, C.
\newblock 2015.
\newblock Wikiqa: A challenge dataset for open-domain question answering.
\newblock In {\em Proceedings of the Conference on EMNLP},  2013--2018.

\bibitem[\protect\citeauthoryear{Yates \bgroup et al\mbox.\egroup
  }{2007}]{yates2007textrunner}
Yates, A.; Cafarella, M.; Banko, M.; Etzioni, O.; Broadhead, M.; and Soderland,
  S.
\newblock 2007.
\newblock Textrunner: open information extraction on the web.
\newblock In {\em The Annual Conference of the NAACL},  25--26.

\bibitem[\protect\citeauthoryear{Yih \bgroup et al\mbox.\egroup
  }{2013}]{Yih2013LCLR}
Yih, W.-t.; Chang, M.-W.; Meek, C.; and Pastusiak, A.
\newblock 2013.
\newblock Question answering using enhanced lexical semantic models.
\newblock In {\em Proceedings of Annual Meeting of the Association for
  Computational Linguistics (ACL)},  1744--1753.

\bibitem[\protect\citeauthoryear{Yih \bgroup et al\mbox.\egroup
  }{2015}]{Yih2015acl}
Yih, W.-t.; Chang, M.-W.; He, X.; and Gao, J.
\newblock 2015.
\newblock Semantic parsing via staged query graph generation: Question
  answering with knowledge base.
\newblock In {\em Proceedings of ACL}.

\bibitem[\protect\citeauthoryear{Yin \bgroup et al\mbox.\egroup
  }{2015}]{yin2015answering}
Yin, P.; Duan, N.; Kao, B.; Bao, J.; and Zhou, M.
\newblock 2015.
\newblock Answering questions with complex semantic constraints on open
  knowledge bases.
\newblock In {\em Proceedings of the 24th ACM International on CIKM},
  1301--1310.
\newblock ACM.

\bibitem[\protect\citeauthoryear{Yin \bgroup et al\mbox.\egroup
  }{2016}]{yin2015abcnn}
Yin, W.; SchÃ¼tze, H.; Xiang, B.; and Zhou, B.
\newblock 2016.
\newblock Abcnn: Attention-based convolutional neural network for modeling
  sentence pairs.
\newblock {\em TACL} 4:259--272.

\bibitem[\protect\citeauthoryear{Yu \bgroup et al\mbox.\egroup
  }{2014}]{yu2014bicnn}
Yu, L.; Hermann, K.~M.; Blunsom, P.; and Pulman, S.
\newblock 2014.
\newblock Deep learning for answer sentence selection.
\newblock {\em NIPS Deep Learning and Representation Learning Workshop}.

\bibitem[\protect\citeauthoryear{Zeiler}{2012}]{zeiler2012adadelta}
Zeiler, M.~D.
\newblock 2012.
\newblock Adadelta: an adaptive learning rate method.
\newblock {\em arXiv preprint arXiv:1212.5701}.

\end{thebibliography}
\end{document}